\documentclass{article}

\usepackage{lscape}  % 横向排版
\usepackage{graphicx}  % 用于表格缩放
\usepackage{colortbl} % 在导言区引入宏包
\usepackage{xcolor}   % 用于定义颜色
\usepackage{multirow} 
\usepackage{amsmath}
\usepackage{float}

    \usepackage[preprint]{neurips_2025}

\usepackage[utf8]{inputenc} % allow utf-8 input
\usepackage[T1]{fontenc}    % use 8-bit T1 fonts
\usepackage{hyperref}       % hyperlinks
\usepackage{url}            % simple URL typesetting
\usepackage{booktabs}       % professional-quality tables
\usepackage{amsfonts}       % blackboard math symbols
\usepackage{nicefrac}       % compact symbols for 1/2, etc.
\usepackage{microtype}      % microtypography
\usepackage{xcolor}         % colors

\title{Squeeze10-LLM: Squeezing LLMs' Weights by 10 Times via a Staged Mixed-Precision Quantization Method}

\author{%
  Qingcheng Zhu\textsuperscript{1}\thanks{Equal contribution.},\quad
  Yangyang Ren\textsuperscript{1}\footnotemark[1],\quad 
  Linlin Yang\textsuperscript{2}\thanks{Corresponding author.}, \quad
  Yanjing Li\textsuperscript{1},\quad 
  Sheng Xu\textsuperscript{1},\quad \\
  Haodong Zhu\textsuperscript{1},\hspace{0pt}
  Juan Zhang\textsuperscript{1},\quad
  Runqi Wang\textsuperscript{3},\quad
  Baochang Zhang\textsuperscript{1}\hspace{0pt}\\
  \textsuperscript{1}Beihang University \quad
  \textsuperscript{2}Communication University of China
  \textsuperscript{3}Beijing Jiaotong University\\
  \texttt{bczhang@buaa.edu.cn}\quad
  \texttt{lyang@cuc.edu.cn}
}

\begin{document}

\maketitle

\begin{abstract}

Deploying large language models (LLMs) is challenging due to their massive parameters and high computational costs. Ultra low-bit quantization can significantly reduce storage and accelerate inference, but extreme compression (i.e., {mean bit-width} $\le 2$) often leads to severe performance degradation.
To address this, we propose Squeeze10-LLM, effectively ``\textbf{squeezing}'' 16-bit LLMs' weights by \textbf{10} times. Specifically, Squeeze10-LLM is a staged mixed-precision post-training quantization (PTQ) framework and achieves an average of 1.6 bits per weight by quantizing 80\% of the weights to 1 bit and 20\% to 4 bits.
We introduce Squeeze10-LLM with two key innovations: Post-Binarization Activation Robustness (PBAR) and Full Information Activation Supervision (FIAS). PBAR is a refined weight significance metric that accounts for the impact of quantization on activations, improving accuracy in low-bit settings. FIAS is a strategy that preserves full activation information during quantization to mitigate cumulative error propagation across layers.
Experiments on LLaMA and LLaMA2 show that Squeeze10-LLM achieves state-of-the-art performance for sub-2bit weight-only quantization, improving average {accuracy} from 43\% to 56\% on six zero-shot classification tasks---a significant boost over existing PTQ methods. Our code will be released upon publication.

\end{abstract}    
\section{Introduction}
\label{sec:intro}

\begin{figure}[t]
    \centering
    \includegraphics[width=0.45\textwidth]{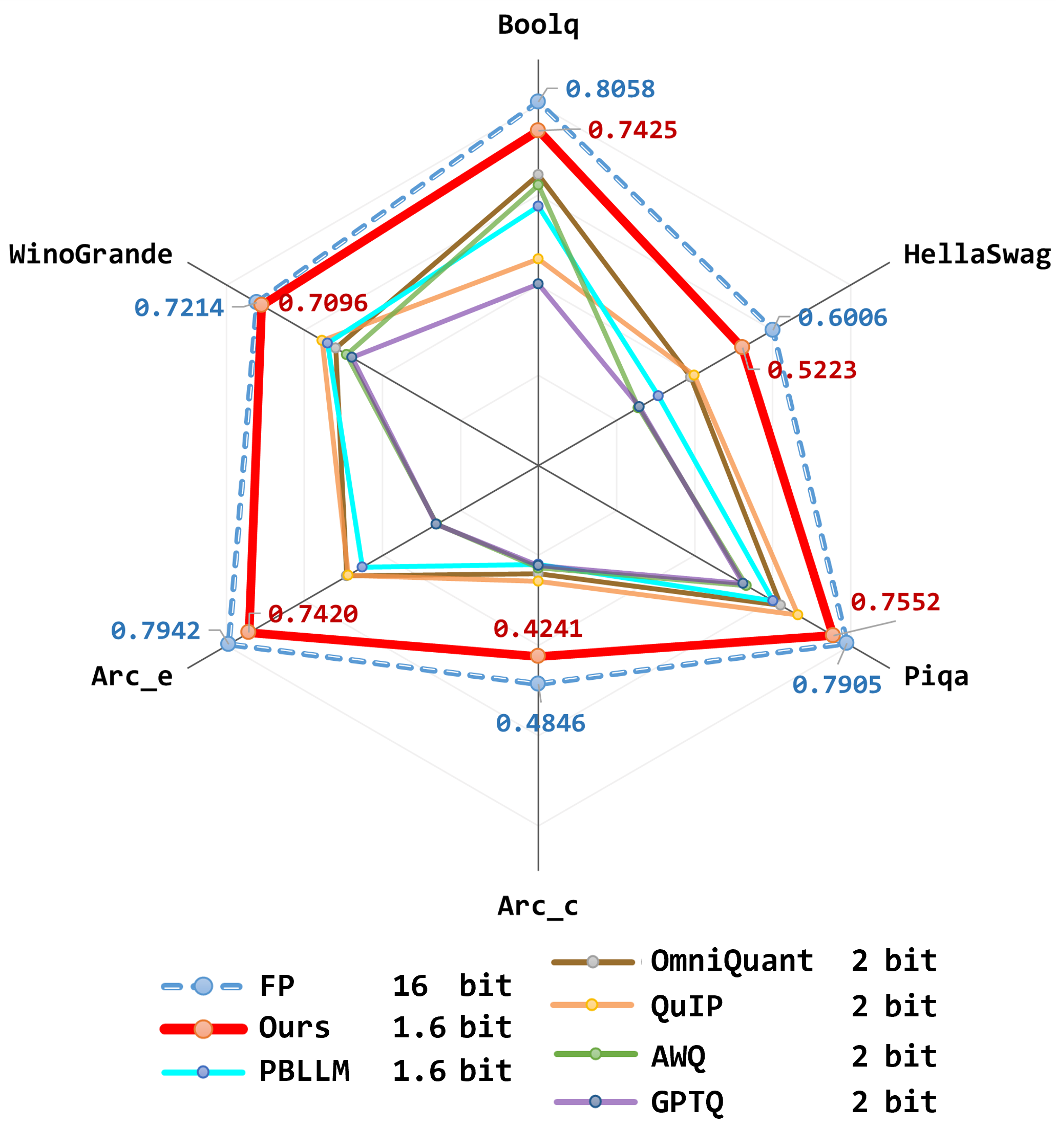}
    \caption{Accuracy comparisons of LLaMA2-13B on 6 zero-shot classification tasks~\cite{clark2019boolq,zellers2019hellaswag,bisk2020piqa,sakaguchi2021winogrande,clark2018think}. Our proposed Squeeze10-LLM (red line) significantly outperforms existing ultra low-bit quantization methods, even comparable to 16-bit full-precision weight (blue dotted line).}
    \label{fig:head}
\end{figure}
In recent years, large language models (LLMs) have gained significant attention in artificial intelligence, driven by the success of models like ChatGPT~\cite{achiam2023gpt,jaech2024openai,hurst2024gpt} and DeepSeek~\cite{lu2024deepseek,liu2024deepseek-v3,liu2024deepseek-v2,guo2025deepseek}. However, as model sizes continue to grow, their massive parameter counts pose significant challenges for deployment on memory-constrained devices. Ultra-low quantization offers a potential solution by drastically reducing storage and computational costs. 
As illustrated in Figure\,\ref{fig:head}, mainstream methods~\cite{frantar2022gptq,lin2024awq,shao2023omniquant,xiao2023smoothquant,chee2023quip} suffer from severe performance degradation in ultra-low-bit settings. Mixed-precision quantization, such as LLM-MQ~\cite{li2023llm}, provides better compression while partially mitigating this degradation. However, even state-of-the-art mixed-precision approaches still exhibit substantial performance gaps compared to models with 16-bit full precision. 

For instance, PB-LLM~\cite{shang2023pb}, as one of the most representative mixed-precision approaches, suffers a 22.8\% accuracy drop compared to its full-precision counterpart, as shown in Figure\,\ref{fig:head}, which greatly affects its effectiveness in real-world applications.
This raises a critical question: Can we further close the performance gap of mixed-precision ultra-low-bit quantization while maintaining efficiency?

Achieving ultra-low-bit quantization while minimizing performance degradation remains a significant challenge. 
% \tr{Existing ultra-low-bit quantization for LLM focus on XXX from the aspects of XXX and XXX, and improve the performance}.
{Existing ultra-low-bit quantization for LLMs focuses on retaining a small fraction of salient weights to improve performance. 
For example, previous methods{~\cite{zheng2024mixllm,huang2024slim,shang2023pb}} rely on output error-based salience metrics, such as global loss functions~\cite{zheng2024mixllm} or information entropy~\cite{huang2024slim}, to determine weights that require higher precision. Recent approaches leverage Hessian-based metrics~{\cite{shang2023pb}}, yet these still derive from output errors.

However, it is still non-trivial to estimate accurate weight salience to help retain critical information.
A critical oversight is that the underutilizaton of activations values, which directly reflect weight contributions.}
Moreover, accumulated quantization errors can degrade the performance of deep networks.
{For quantization errors}, post-training quantization (PTQ) relies on activation values to compute Hessians, but as quantization progresses, the distribution of activations shift layer by layer, leading to cumulative errors. These shifts become severer in ultra-low-bit settings, degrading performance in later layers.

To address the aforementioned challenges, we propose Squeeze10-LLM, effectively ``\textbf{squeezing}'' 16-bit LLMs' weights by \textbf{10} times. Specifically, Squeeze10-LLM is a staged mixed-precision post-training quantization framework that achieves 1.6-bit weight-only quantization by binarizing 80\% of weights while retaining 4-bit precision for the remaining 20\%, effectively reducing the original 16-bit representation to an average of 1.6 bits per weight. Squeeze10-LLM incorporates two key techniques, i.e., Post-Binarization Activation Robustness (PBAR) and Full Information Activation Supervision (FIAS), to {achieve accurate weight salience estimation, reduce quantization error accumulation}, and {therefore enhance quantization efficiency and accuracy}.
{For PBAR}, we introduce an activation-aware metric that considers the impact of binarization on activation range. We identify weights that significantly expand post-binarization activation ranges and upgrade their importance in the salience ranking. This prevents unnecessary precision allocation to less critical weights, ensuring better retention of key information.
{For FIAS}, we consistently use original pretrained activations when computing Hessians, rather than updating activations layer by layer. This prevents the accumulation of activation shifts and maintains stable quantization quality across all layers, particularly under extreme compression ratios.
In summary, the contributions of
our work are as follows:
\begin{itemize}
    \item We propose Squeeze10-LLM, a staged mixed-precision PTQ method that achieves 1.6-bit weight-only quantization by binarizing 80\% of weights and preserving 4-bit precision for the remaining 20\%.

    \item We introduce PBAR, a novel salience metric that improves weight selection by considering activation range changes after binarization.

    \item We propose FIAS to utilize original activations to supervise PTQ , which prevents activation shifts and ensures stable and efficient quantization.

    \item Squeeze10-LLM achieves 10$\times$ compression of 16-bit pretrained weights with minimal performance degradation (see Figure\,\ref{fig:head}). Evaluated on LLaMA and LLaMA2 models, it establishes state-of-the-art (SOTA) results among all PTQ methods at 2-bit and below.  %, demonstrating significant improvements in accuracy.
\end{itemize}

\section{Related Works}

\subsection{Uniform-Precision Quantization for LLMs}
%Quantization, as a model compression technique, can significantly reduce the storage and computational costs of deep learning models. In recent years, many studies have applied it to LLMs and achieved promising results. In this paper, we focus exclusively on post-training quantization (PTQ).

Uniform-precision quantization compresses the weights or activations of a neural network to a lower-bit width. % (\eg, 1 bit, 2 bit, or 4 bit).
For LLMs, uniform-precision quantization commonly opts for post-training quantization (PTQ). According to the quantization targets, it can be  divided into weight-only quantization and weight-activation quantization.

% Quantization serves as a pivotal compression technique, well reducing both storage and computational overhead. Recent advancements have demonstrated promising applications of quantization in LLMs, with this paper specifically focusing on post-training quantization (PTQ).

%LLM quantization techniques can be broadly classified into two categories: weight-only quantization and weight-activation quantization. The former focuses on reducing model storage, while the latter also aims to accelerate inference. 
%
%For weight-only quantization, GPTQ~\cite{frantar2022gptq} improves OBQ~\cite{frantar2022optimal} and introduces layer-wise quantization, which leverages second-order information to compensate for quantization errors. QuIP~\cite{chee2023quip} performs quantization with incoherence processing, enhancing weight distribution and enabling effective 2-bit quantization. AWQ~\cite{lin2024awq} and OWQ~\cite{lee2024owq} point out that during weight quantization, it is essential to consider the impact of activation outliers on weights.
%
In the context of weight-only quantization, which primarily focuses on reducing model storage, GPTQ~\cite{frantar2022gptq} enhances OBQ~\cite{frantar2022optimal} by introducing layer-wise quantization and leveraging second-order information to compensate for quantization errors. QuIP~\cite{chee2023quip} further advances 2-bit quantization by adjusting weight distribution, while AWQ~\cite{lin2024awq} and OWQ~\cite{lee2024owq} emphasize the necessity of considering activation outliers, ensuring greater robustness in the quantization process.
%
%For weight-activation quantization, SmoothQuant~\cite{xiao2023smoothquant} and Outlier Suppression~\cite{wei2022outlier} addresses the outlier issue by applying per-channel scaling transformations, converting the challenge of activation quantization into a weight quantization problem. OmniQuant~\cite{shao2023omniquant} optimizes quantization with learnable clipping and equivalent transformation. ZeroQuant~\cite{yao2022zeroquant} and RPTQ~\cite{yuan2023rptq} perform grouped quantization using finer granularity and clustering methods. In this study, we focus primarily on weight quantization, aiming to achieve high-compression-ratio binarization combined with high-bit mixed quantization.
For weight-activation quantization, it reduces model size and accelerates inference, addressing outlier management through various optimization techniques. SmoothQuant~\cite{xiao2023smoothquant} and Outlier Suppression~\cite{wei2022outlier} mitigate the impact of activation outliers by employing per-channel scaling transformations, effectively transforming activation quantization into a weight quantization problem. Building on this, OmniQuant~\cite{shao2023omniquant} introduces learnable clipping and equivalent transformation to further enhance quantization efficiency. Meanwhile, ZeroQuant~\cite{yao2022zeroquant} and RPTQ~\cite{yuan2023rptq} refine granularity control by leveraging grouped quantization and clustering methods, improving {accuracy} while maintaining computational efficiency.

%This study primarily focuses on weight quantization, striving to achieve high compression ratios through binarization while preserving model performance via mixed-precision quantization strategies.

\subsection{Mixed-Precision Quantization for LLMs}

The key to leveraging mixed-precision quantization lies in accurately assessing salient weights and judiciously allocating bit-widths. 
{For salient weights}, LLM-MQ~\cite{li2023llm} applies ultra-low precision for normal weights, while preserving outliers in FP16 precision, optimizing the model's efficiency. 
MixLLM~\cite{zheng2024mixllm} adopts a global loss function-based evaluation approach, identifying critical features across the model and assigning higher bit-widths to those with greater significance. 
{For bit-widths}, SILM-LLM~\cite{huang2024slim} allocates bit-widths by minimizing information entropy disparities between the quantized and original weights, ensuring optimal precision for different weight groups. The follow-up PMPD~\cite{chen2024progressive} further refines precision allocation by adjusting the model’s bit-widths between the prefilling and decoding stages, optimizing performance dynamically across varying sequence lengths. 
APTQ~\cite{guan2024aptq} enhances layer-wise precision allocation by calculating the average trace of the Hessian matrix. Similarly, AMLQ~\cite{ou2024adaptive} employs a search-based method to determine the most efficient mixed-precision configuration, minimizing output error. Although these advancements have well improved quantization, performance typically deteriorates sharply below 2 bits. 

%We herein introduce a staged mixed-precision PTQ method that introduces a novel weight salience measure and achieves 1.6-bit weight-only quantization with impressive performance.

\label{sec:relate}

\section{Preliminaries}
\label{sec:pre}
%Quantization, sparsification, and knowledge distillation are the basic methods for compressing large language models. 
%
%In this paper, we give a mixed-precision quantization scheme for weights, in which the quantization of high and low bit-widths is achieved by general uniform quantization and binarization, respectively.

To achieve mixed-precision quantization for model weights, high- and low-bit-width quantization can be accomplished through standard uniform quantization and binarization, respectively.
%For $k$-bit uniform asymmetric quantization, the full precision weights \(w\) will be quantized as \(w_q\) and then de-quantized as \(\hat{w}\) by the following equations:
%
For a $k$-bit uniform asymmetric quantization, a full-precision weight $w$ is quantized to $w_q$ and then dequantized to $\hat{w}$ using the following equations:
\begin{equation}
\begin{aligned}
    w_q &= \mathrm{clamp}\left(\left\lfloor\frac{w}{s}\right\rceil + z,0,2^{k}-1\right), \\
    \hat{w} &= s \times (w_q - z),
\end{aligned}
\end{equation}
where \(s=\frac{max(w)-min(w)}{2^k-1}\) and \(z=\left\lfloor\frac{-min(w)}{s}\right\rceil\) are the scaling factor and zero point, respectively. The function \(\text{clamp}(\cdot)\) ensures that values remain within a specified range and is defined as:
\begin{equation}
\mathrm{clamp}(x,a,b) = 
\begin{alignedat}{2}
&\begin{cases}
a, & x \leq a, \\
x, & a < x < b, \\
b, & x \geq b.
\end{cases}
\end{alignedat}
\end{equation}

%Similarly, binarization is to compress the model weights from full precision to 1bit, and this can be achieved by applying sign function.
% \begin{figure*}[htbp]
\begin{figure}[htbp]
% \begin{figure}[H]
    \centering
    \includegraphics[width=1.0\textwidth,height=0.3\textheight]{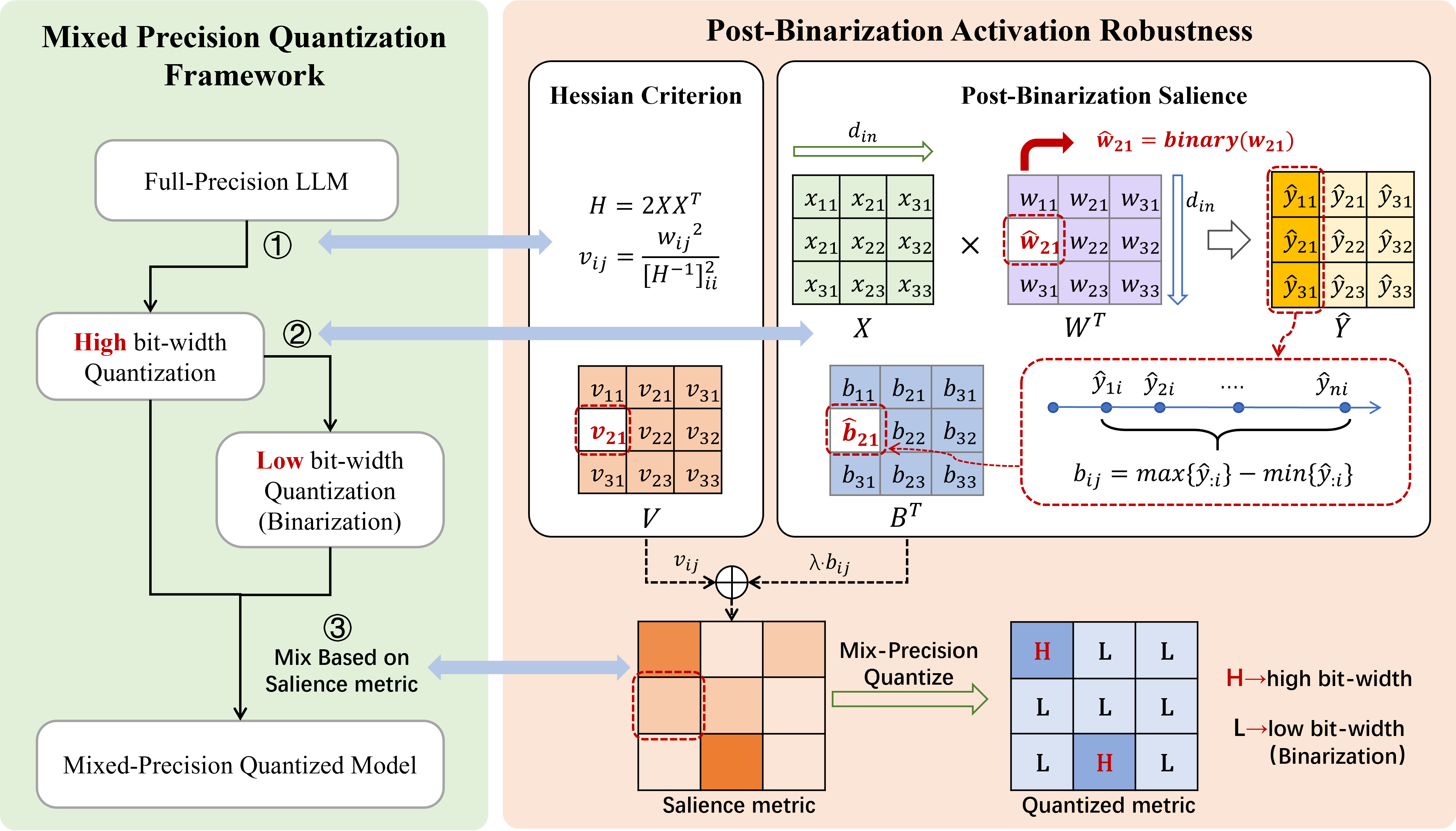}
    \caption{Overview of Squeeze10-LLM. Squeeze10-LLM is the Mixed-Precision Quantization Framework with stepwise low-bit quantization (see Sec.\,\ref{sec:framework}). Especially, it leverages the Post-Binarization Activation Robustness metric to represent the salient weights (see Sec.\,\ref{sec:pbar}), and Information Activation Supervision to minimize layer-wise accumulated errors (see Figure\,\ref{fig:FIAS} and Sec.\,\ref{sec:fias}).}
    % Combine the Hessian Criterion and Post-Binarization Activation Salience to represent the salience matrix of weights.}
    \label{fig:framework}
\end{figure}
% \end{figure*}

Similarly, to achieve 1-bit quantization, binarization is applied to model weights using the sign function
\begin{equation}
\mathrm{sign}(x) = \begin{alignedat}{2}
&\begin{cases}
-1, & x \leq 0, \\
1, & x > 0.
\end{cases}
\end{alignedat}
\end{equation}

% This approach effectively reduces the computational and memory requirements of LLMs while maintaining performance through mixed-precision quantization.
% \section{Methodology}
% \label{sec:method}

%\section{Partial Binary Quantization Framework}
\section{Squeeze10-LLM}
\label{sec:method}

%In this section, we propose a 4-bit-based partial binarization framework, which demonstrates significant advantages in ultra low-bit quantization. First, we introduce a staged mixed-bit quantization framework. Then, we apply the Sparse Min-Max quantization formula to perform high-bit quantization on all weights. Based on this, we further binarize the non-significant weights according to Post-Binarization Activation Robustness (PBAR). During the quantization process, we also leverage Full Information Activation Supervision (FIAS) to supervise the quantization process layer by layer, minimizing quantization error.

In this section, we introduce Squeeze10-LLM, a staged mixed-precision quantization framework designed to push the boundaries of ultra-low-bit quantization. Our method strategically balances precision and efficiency by leveraging a staged mixed-bit quantization approach.

%
%To achieve this, we first apply a sparse uniform quantization scheme to perform high-bit quantization on all models weights, ensuring optimal numerical representation. Building upon this foundation, we selectively binarize non-significant weights using our Post-Binarization Activation Robustness (PBAR) criterion, which identifies weights that can be safely compressed without degrading model performance.
%
%Furthermore, to mitigate quantization-induced information loss, we introduce Full Information Activation Supervision (FIAS)---a layer-wise guidance mechanism that supervises the quantization process, effectively minimizing errors and preserving activation fidelity.

%By integrating these techniques, Squeeze10-LLM establishes a robust and efficient quantization paradigm, unlocking new potential for ultra-low-bit LLMs.

% \input{figures/Multi-Bit-Compare}
% \input{figures/FIAS}
\subsection{Staged Mixed-Precision Quantization Framework}
\label{sec:framework}
%We propose a staged mixed-bit quantization framework that significantly enhances ultra-low-bit quantization performance.
%
%For low-bit quantization in the mixed-bit quantization, directly applying ultra-low-bit quantization (such as 1-bit or 2-bit) to the original weights leads to a significant decline in model performance. In contrast, if the original weights are first quantized to a higher bit width as an intermediate buffering stage, and then further quantized to lower bits, the performance degradation can be mitigated. Moreover, among the higher-bit quantization levels ranging from 2-bit to 8-bit, we select 4-bit, as Appendix \ref{sec:Appendix} reports that 4-bit outperforms other bit widths, even though higher-bit quantization theoretically preserves more information.
%
A key quantization challenge is the severe accuracy degradation that occurs when directly applying ultra-low-bit quantization (eg 1-bit) to full-precision weights. To mitigate this issue, we adopt a two-stage strategy: rather than binarizing weights outright, we first quantize them to a higher bit width as an intermediate buffering stage before applying further quantization. This staged process helps preserve essential information and stabilize performance. Furthermore, we adopt 4-bit precision for higher bit-width settings, with detailed discussions provided in Appendix \ref{appendix-high-bit} and Section \ref{subsec:analysis}.

% For intermediate bits, possible bits range from 2-bit to 8-bit. Figure\,\ref{fig:Multi-Bit-Compare} demonstrates that the activation distribution obtained using 4-bit quantization is the closest to that of the full precision (FP) model. The advantage of 4-bit as an intermediate representation is also confirmed by our experiments {(see Sec.\,\ref{subsec:analysis})}. Thus, we use 4-bit quantization. 

Our partial binarization framework consists of three key steps:

(1) \textbf{4-bit Uniform Quantization:} We first apply 4-bit uniform quantization to introduce sparsity into the original weights, serving as an intermediate step before lower-bit quantization (see Sec.\,\ref{sec:pre}).

(2) \textbf{Salience-Based Binarization with PBAR:} Building upon the 4-bit quantized weights, we compute the Post-Binarization Activation Robustness (PBAR) metric, considering Hessian with weight outliers and post-binarization activation salience. This metric enables selective binarization of non-salient weights, ensuring safe compression with less degrading model performance (see Sec.\,\ref{sec:pbar}).

(3) \textbf{Mixed-Bit Supervision with FIAS:} To further mitigate quantization-induced information loss, we introduce Full Information Activation Supervision (FIAS)---a layer-wise guidance mechanism that supervises the quantization process, effectively minimizing errors and preserving activation distributions (see Sec.\,\ref{sec:fias}). 

By integrating these techniques, Squeeze10-LLM establishes a robust and efficient quantization paradigm, unlocking new potential for ultra-low-bit LLMs.

\subsection{Salience-Based Binarization with PBAR\label{sec:pbar}}

\noindent\textbf{Hessian with Weight Outliers}.
{When an LLM performs forward propagation in a linear layer, the output $\mathbf{Y}$ of the layer can be calculated based on the input activations $\mathbf{X} \in \mathbb{R}^{N\times d_{in}}$, and the weights $\mathbf{W} \in \mathbb{R}^{d_{out}\times d_{in}}$.
% the input activations are denoted as $\mathbf{X} \in \mathbb{R}^{N\times d_{in}}$ and the weights as $\mathbf{W} \in \mathbb{R}^{d_{out}\times d_{in}}$. 
Here, $N,d_{in}, d_{out}$ are the number of tokens, input dimensions, and output dimensions respectively. The output $\mathbf{Y}$ of the layer can be written as:}
%
% In the forward propagation of a linear layer in LLM, the input activations are denoted as $\mathbf{X} \in \mathbb{R}^{N\times d_{in}}$, while the weight matrix is represented as  $\mathbf{W} \in \mathbb{R}^{d_{out}\times d_{in}}$, where $N,d_{in}, d_{out}$ correspond to the number of tokens, input dimensions, and output dimensions, respectively. }Consequently, the output $\mathbf{Y}$ of the layer is computed as follows:
%
\begin{equation}
    \begin{aligned}
\label{equation:forward}
    \mathbf{Y} = \mathbf{XW}^\mathrm{T}  
    \text{~~~where } \mathbf{Y}_{ij} = \sum_{k=1}^{c_{in}} \mathbf{X}_{ik} \cdot \mathbf{W}_{jk}.
\end{aligned}
\end{equation}

According to SparseGPT~\cite{frantar2023sparsegpt}, we define a salient matrix $\mathbf{V}$ based on Hessian criterion and each element $v_{ij}$ in it can be calculated as follows: 
\begin{equation}
    \mathbf{v}_{ij} = \frac{\mathbf{w}_{ij}^{2}}{[\mathbf{H}^{-1}]_{ii}^{2}},
    \label{importance measure}
\end{equation}
where \(\mathbf{H}\) is the Hessian matrix of the quantization loss function in GPTQ~\cite{frantar2022gptq}, which serves as a criterion for detecting significant weights. The Hessian can be derived as the product of the activation metric matrix \(\mathbf{X}\) and its transpose, scaled by a factor of 2:
\begin{equation}
    \mathbf{H} = 2\mathbf{X} \mathbf{X}^T.
\end{equation}

The salience matrix tends to preserve elements with larger absolute values, incorporating information related to the inverse Hessian of diagonal elements (\emph{i.e.}, the magnitude of input activations). However, as seen in Eq.\,(\ref{equation:forward}), the salience matrix does not directly capture the activation information from the output, which is crucial for the assessment of salience. To address this, below, we introduce an enhanced saliency metric that accounts for activation range, effectively capturing variations before and after quantization.

\noindent\textbf{Post-Binarization Activation Salience}.
To better quantify the influence of individual weights on output activations, we propose a measurement metric that integrates the change in activation range for each channel following binarization, inspired by JSQ~\cite{guo2024compressing}. Since the $j$-th output channel is determined solely by the $j$-th row of weight matrix $\mathbf{W}$, we define a post-binarization activation salience matrix $\mathbf{B} \in \mathbb{R}^{d_{out} \times d_{in}}$ as follows:
%
%We establish a measurement metric to capture the influence of each weight on the output activation. We incorporate the change in activation range for each channel after binarizing an individual weight as part of the weight-salidence metric. Since the $j$th output channel is solely dependent on the $j$th row of the weight matrix $W$, we define a measurement matrix $B \in \mathbb{R}^{d_{out} \times d_{in}}$ based on this insight:
\begin{equation}
    \begin{aligned}
    &\mathbf{B}_{ij} =\|\mathbf{\mathbf{\hat{Y}}}_{:,i}\|_{\infty}-\|\mathbf{\mathbf{\hat{Y}}}_{:,i}\|_{min},\\
    &\text{where~~} \mathbf{\hat{Y}} = \mathbf{X} \cdot (\mathcal{Q}(\mathbf{W}; i; j))^T.
\end{aligned}
\end{equation}

Here, $\mathcal{Q}(\mathbf{W}; i; j)$ represents a measurement matrix that quantize the element at $i$-th row and $j$-th column from high-bit precision to 1-bit. Each entry in $\mathbf{B}$ reflects a crucial property: the extent to which the activation range changes when an individual weight is quantized. By leveraging this metric, we can determine whether a given weight should be binarized based on its impact on the activation range.

\noindent\textbf{Post-Binarization Activation Robustness (PBAR)}.
By combining the Hessian-based saliency metric with weight outlier detection and post-binarization activation range analysis, we derive the final salience metric $\mathbf{M} \in \mathbb{R}^{d_{out} \times d_{in}}$ for mixed-precision quantization:
\begin{equation}
\mathbf{M}=\mathbf{V}+\lambda\mathbf{B},
    \label{Salience metric}
\end{equation}
where $\lambda$ is a scaling factor that balances the contributions of the two salience metrics.

\begin{figure}[htbp]
    \centering
    \includegraphics[width=\textwidth]{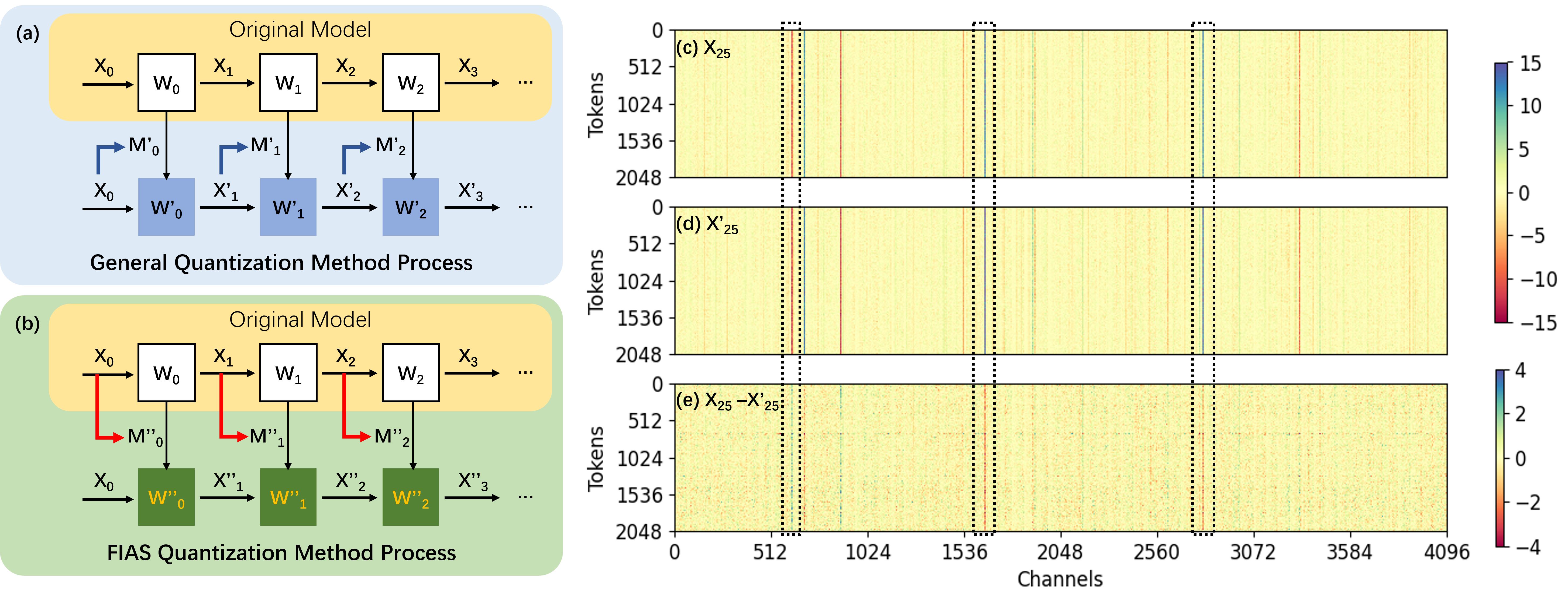}
    \caption{Comparison of the structural diagrams of Full Information Activation Supervision (FIAS) and visualization of certain activation values. (a) and (b) present the comparison of the quantization processes between General and FIAS. \(X\), \(W\), and \(M\) represent the activation values, weights, and salience measurements respectively. The superscripts \('\) and \(''\) denote the intermediate quantities obtained by two different methods, and the subscript numbers indicate the sequence numbers of the network structures. {For the same input, FIAS employs the activation values of the original model for supervision, and this can reduce the weight quantization shift.} In (c) and (d), the activation value outputs of the key projection structure in the 25-th layer of LLaMA-7B are shown before and after the utilization of the general quantization method. Figure (e) presents the difference of these two cases. {Some channels are highlighted by the dashed boxes, and it is clear that quantization can actually lead to significant numerical shifts in the activation values.}}
    \label{fig:FIAS}
\end{figure}

The metric defined in Eq.\,(\ref{Salience metric}) follows an intuitive design principle:
If a weight has a large absolute value or its binarization significantly alters the activation range, it should be retained with higher precision to preserve information. Conversely, weights that contribute minimally to the activation range can be safely quantized to 1-bit. By utilizing this PBAR metric, we achieve a superior trade-off---preserving outliers when necessary for information retention while mitigating their adverse effets on quantization. This approach enhances overall quantization efficiency and maintains model robustness.

\subsection{Mixed-Bit Supervision with FIAS\label{sec:fias}}

% \input{figures/FIAS}
%Based on the above, activation matrices play the role of measuring the importance of weights. As depicted in Figure\,\ref{fig:FPAS}, when quantization of a particular layer is completed, the value change in weights triggers an update of the output of that layer, which is also activation of the next layer. Using the updated activation to drive the quantization of the next layer seems to be a general and tacitly accepted practice in general method, but we find that this is not optimal in the case of high-fold quantization. Therefore, a better supervision is urgent to improve this activation.

In the context of quantization, activation matrices play a crucial role in measuring the salient weights. As illustrated in Figure\,\ref{fig:FIAS}(a), once a particular layer undergoes quantization, changes in its weight values inevitably lead to modifications in the layer's output, which subsequently serves as the activation input for the next layer. A widely accepted yet implicit practice in conventional quantization methods is to use these updated activations to guide the quantization of subsequent layers. However, we find that this approach is suboptimal in scenarios involving aggressive quantization with high bit-reduction ratios. Thus, an effective supervision mechanism is essential to enhance the quality of the activation and improve the performance of the quantization.

\begin{table*}[!t]
\caption{Performance comparison of the LLaMA2 family across different quantization methods on three text-generation tasks and six zero-shot classification tasks. The gray-marked parts represent the performance of the pre-trained model, while the red-marked and blue-marked parts indicate the best and second-best performance among quantization methods, respectively.}
\label{Llama2 result}
%\vskip -0.1in
% \begin{center}
\renewcommand{\arraystretch}{1.2} % 调整行高
\begin{small}
\resizebox{\linewidth}{!}{  % 使表格宽度适应页面横宽
\begin{tabular}{ccc|ccc|cccccc|c}
\toprule
% \cline{4-10}  % 在accuracy下画线
% \multicolumn{1}{c}{\textbf{Model}} & 
% \multicolumn{1}{c}{\textbf{Method}} & 
% \multicolumn{1}{c}{\textbf{WBits}} &
\multirow{2}{*}{\textbf{Model}} & 
\multirow{2}{*}{\textbf{Method}} & 
\multirow{2}{*}{\textbf{\#W-Bits}} & 
\multicolumn{3}{c}{\textbf{Perplexity$\downarrow$}} &
\multicolumn{7}{c}{\textbf{Accuracy(\%)$\uparrow$}} \\
\cline{4-6} \cline{7-13}
&&& \multicolumn{1}{c}{WikiText2} & 
\multicolumn{1}{c}{Ptb} & 
\multicolumn{1}{c}{C4} &
\multicolumn{1}{c}{BoolQ} & 
\multicolumn{1}{c}{HellaSwag} & 
\multicolumn{1}{c}{PIQA} & 
\multicolumn{1}{c}{WinoGrande} & 
\multicolumn{1}{c}{ARC-c} & 
\multicolumn{1}{c}{ARC-e} & 
\multicolumn{1}{c}{Avg.} \\
% \addlinespace[1ex]
\midrule
\multirow{7}{*}{LLaMA2-7B} 
& \cellcolor{gray!30}FP  &\cellcolor{gray!30} 16  &\cellcolor{gray!30}5.47&\cellcolor{gray!30}37.91&\cellcolor{gray!30}7.26& \cellcolor{gray!30}77.74 & \cellcolor{gray!30}57.13 & \cellcolor{gray!30}78.07 & \cellcolor{gray!30}69.22 & \cellcolor{gray!30}43.52 & \cellcolor{gray!30}76.35 & \cellcolor{gray!30}64.86 \\
& GPTQ  & 2 &1.99e3&3.65e4&4.13e3& 41.04 & 26.01 & 51.96 & 49.09 & 21.33 & 25.55 & 34.79 \\
& AWQ  & 2 & 2.23e5&2.02e5&1.69e5&62.17&25.59&53.32&49.17&\cellcolor{blue!10}22.78&26.14&39.86\\
& QuIP  & 2 &98.33&1.03e3&83.87& 54.59 & 28.32 & 54.95 & 52.80 & 19.62 & 32.28 & 37.59 \\
& OmniQuant  & 2 &54.13&\cellcolor{blue!10}822.19&130.86& 57.06 & 29.01 & 55.55 & 51.22 & 20.82 & 31.73 & 37.67 \\
& PB-LLM  & 1.6 &\cellcolor{blue!10}12.29&5.74e3&\cellcolor{blue!10}26.03&\cellcolor{blue!10}63.79 & \cellcolor{blue!10}34.33 &\cellcolor{blue!10} 61.10 & \cellcolor{blue!10}56.43 & 22.18 & \cellcolor{blue!10}45.71 & \cellcolor{blue!10}43.95 \\
& Squeeze10-LLM  & 1.6 &\cellcolor{red!10}9.96&\cellcolor{red!10}409.62&\cellcolor{red!10}12.8&\cellcolor{red!10}67.43 & \cellcolor{red!10}46.03 & \cellcolor{red!10}72.20 & \cellcolor{red!10}64.56 & \cellcolor{red!10}32.94 & \cellcolor{red!10}64.48 & \cellcolor{red!10}\textbf{56.04} \\

\midrule
\multirow{7}{*}{LLaMA2-13B} & \cellcolor{gray!30}FP  & \cellcolor{gray!30}16 &\cellcolor{gray!30}4.88&\cellcolor{gray!30}50.94&\cellcolor{gray!30}6.73& \cellcolor{gray!30}80.58&\cellcolor{gray!30}60.06&\cellcolor{gray!30}79.05&\cellcolor{gray!30}72.14&\cellcolor{gray!30}48.46&\cellcolor{gray!30}79.42&\cellcolor{gray!30}67.83 \\
                          & GPTQ  & 2 &306.08&4.31e3&1.22e3& 40.24&25.85&52.39&47.83&22.27&26.18&35.79\\
                          
                          & AWQ  & 2  & 1.22e5&1.14e5&9.56e4&62.17&25.59&53.32&49.17&22.78&26.14&39.86\\
                          & QuIP  & 2  & \cellcolor{blue!10}13.93&377.29&\cellcolor{blue!10}14.36&45.75&\cellcolor{blue!10}39.89&\cellcolor{blue!10}66.43&\cellcolor{blue!10}55.41&\cellcolor{blue!10}25.68&48.78&46.99\\
                          & OmniQuant  & 2 &19.69&814.69&30.14& \cellcolor{blue!10}64.43&39.06&62.08&52.01&24.06&\cellcolor{blue!10}49.16&\cellcolor{blue!10}48.47\\
                          & PB-LLM  & 1.6 & 26.19&\cellcolor{blue!10}369.56&55.27&57.49&30.70&60.07&54.06&22.01&45.16&44.92\\
                          & Squeeze10-LLM  & 1.6 &\cellcolor{red!10}7.37&\cellcolor{red!10}170.36&\cellcolor{red!10}10.24&\cellcolor{red!10}74.25&\cellcolor{red!10}52.23&\cellcolor{red!10}75.52&\cellcolor{red!10}70.96&\cellcolor{red!10}42.41&\cellcolor{red!10}74.20&\cellcolor{red!10}\textbf{64.93}\\
\midrule
\multirow{7}{*}{LLaMA2-70B} & \cellcolor{gray!30}FP  & \cellcolor{gray!30}16  & \cellcolor{gray!30}3.32&\cellcolor{gray!30}24.25&\cellcolor{gray!30}5.71&\cellcolor{gray!30}83.79&\cellcolor{gray!30}64.77&\cellcolor{gray!30}82.21&\cellcolor{gray!30}77.90&\cellcolor{gray!30}54.35&\cellcolor{gray!30}82.74 &\cellcolor{gray!30}74.29 \\
                          & GPTQ  & 16  & 46.08&2.27e3&232.48&38.17&26.05&53.97&49.64&21.16&25.84&35.81\\
                          
                          & AWQ  & 2  & 7.25e4&8.06e4&6.57e4&62.17&25.34&52.50&49.49&22.35&25.76&39.6\\
                          & QuIP  & 2  & 9.08&\cellcolor{blue!10}44.58&11.6&64.71&43.42&70.08&61.72&29.69&63.34&55.49\\
                           & OmniQuant  & 2 &6.11&--&\cellcolor{blue!10}7.89& 74.77&\cellcolor{blue!10}56.59&\cellcolor{blue!10}77.20&69.77&40.70&74.20&65.54\\
                          & PB-LLM  & 2 & \cellcolor{blue!10}5.84&47.12&11.36&\cellcolor{blue!10}76.70&53.74&75.03&\cellcolor{blue!10}75.06&\cellcolor{blue!10}48.04&\cellcolor{blue!10}77.74&\cellcolor{blue!10}67.72\\
                          & Squeeze10-LLM  & 1.6 &\cellcolor{red!10}4.74&\cellcolor{red!10}28.31&\cellcolor{red!10}7.11&\cellcolor{red!10}80.76&\cellcolor{red!10}60.02&\cellcolor{red!10}79.33&\cellcolor{red!10}76.8&\cellcolor{red!10}49.06&\cellcolor{red!10}79.00&\cellcolor{red!10}\textbf{70.83}\\

\bottomrule
\end{tabular}
}
\end{small}
% \end{center}
% \vskip -0.1in
\end{table*}
\noindent\textbf{Full Information Activation Supervision (FIAS)}.
%
%As the model weights are quantized, the information contained within the LLMs is subsequently reduced. From the first layer, bias inevitably occurs in the output and becomes increasingly severe with forward computation. The shifted activation lacks sufficient supervisory effect on the quantization of the weights and may even provide misdirection guidance, which is particularly evident in the context of very low bit quantization. Figure \ref{fig:act} illustrates the distribution of model output values for the self-attentive part of the key projection in 7th layer for the LLaMA1-7B model with the same input. Despite the absence of a substantial alteration in the activation values' distribution (\emph{i.e.}, the relative magnitude of the values) across the channel dimensions following quantization of the weights, a more pronounced shift in the range of numerical expression was observed. Specifically, larger fluctuations in values were evident for certain channels within the gray boxes depicted in the figure. Consequently, the activation values' outliers increased in magnitude and became more distant from the prevailing values.
%
As model weights undergo quantization, the amount of preserved information in LLMs diminishes. From the very first layer, quantization-induced biases accumulate progressively throughout forward propagation, exacerbating  distortions in subsequent layers. These shifted activations fail to provide reliable supervision for weight quantization and may even mislead the process, a problem that becomes particularly pronounced under extremely low-bit quantization.
Figure\,\ref{fig:FIAS}(c)-(e) presents a heatmap visualization of activation output in the key projection part within the 7-th layer of the LLaMA-7B model under the same input conditions.
While the relative magnitudes of activation values across channel dimensions remain largely consistent before and after weight quantization, a significant numerical shift is observed. In particular, certain channels {(highlighted within the dashed boxes)} exhibit larger fluctuations, leading to an increase in activation outliers that deviate substantially from the predominant value range.

%Since activation indicates weight importance, it can be said that the quantization process is supervised by the information implied by the activation. FIAS method preserves all original activations (Figure\,\ref{fig:FPAS}), always supervising the quantization process with full-information activations. This approach circumvents the repercussions engendered by fluctuations in activation values during the calculation of the weight importance metric, which will ultimately enhances the performance of model quantization.

% \input{tabs/llama2}

Since activations inherently indicate weight salience, they serve as an implicit supervisory signal for the quantization process. Compared to general methods in Figure\,\ref{fig:FIAS}(a), the FIAS method in Figure\,\ref{fig:FIAS}(b) utilizes the same calculation equations but preserves all original activations throughout quantization, ensuring that full-information activations consistently guide the process. By doing so, FIAS mitigates the distortions caused by fluctuating activation values when computing salient weights. This enhances model quantization performance by preventing misleading supervisory effects and ensuring a more stable optimization trajectory.

\section{Experiments}
\label{sec:exp}

\subsection{Models and Datasets}
We conducted comprehensive experiments on the LLaMA~\cite{touvron2023llama1} and LLaMA2~\cite{touvron2023llama2} model families. To rigorously assess the efficacy of our Squeeze10-LLM, we evaluate perplexity on language generation benchmarks, including WikiText2~\cite{merity2016pointer}, C4~\cite{raffel2020exploring}, and PTB~\cite{marcus-etal-1993-building}, while measuring accuracy on zero-shot reasoning tasks such as PIQA~\cite{bisk2020piqa}, ARC~\cite{clark2018think}, BoolQ~\cite{clark2019boolq}, HellaSwag~\cite{zellers2019hellaswag}, and WinoGrande~\cite{sakaguchi2021winogrande}.

\subsection{Settings}
We benchmarked our method against state-of-the-art quantization methods, including GPTQ~\cite{frantar2022gptq}, AWQ~\cite{lin2024awq}, PBLLM~\cite{shang2023pb}, QuIP~\cite{chee2023quip}, and OmniQuant~\cite{shao2023omniquant}. In our proposed approach, we quantize 20\% of the most salient weights to 4 bits while binarizing the remaining 80\%, which are deemed less critical. For quantization methods that do not support mixed precision, we standardize the bit-width to 2 bits. To ensure a fair comparison and maintain a consistent average bit-width across LLMs, we implement the partially binarized PB-LLM under the same configuration. For the 7B and 13B models, we utilize a single 80G A800 GPU, while for the 30B and 70B models, we employ four 80G A800 GPUs to conduct quantization.

\begin{table*}[!t]
\caption{Performance comparison of the LLaMA family across different quantization methods on three text-generation tasks and six zero-shot classification tasks. The gray-marked parts represent the performance of the pre-trained model, while the red-marked and blue-marked parts indicate the best and second-best performance among quantization methods, respectively.}
\label{Llama1 result}
%\vskip -0.1in
% \begin{center}
\renewcommand{\arraystretch}{1.2} % 调整行高
\begin{small}
\resizebox{\linewidth}{!}{  % 使表格宽度适应页面横宽
\begin{tabular}{ccc|ccc|cccccc|c}
\toprule
% \cline{4-10}  % 在accuracy下画线
% \multicolumn{1}{c}{\textbf{Model}} & 
% \multicolumn{1}{c}{\textbf{Method}} & 
% \multicolumn{1}{c}{\textbf{WBits}} &
\multirow{2}{*}{\textbf{Model}} & 
\multirow{2}{*}{\textbf{Method}} & 
\multirow{2}{*}{\textbf{\#W-Bits}} & 
\multicolumn{3}{c}{\textbf{Perplexity$\downarrow$}} &
\multicolumn{7}{c}{\textbf{Accuracy(\%)$\uparrow$}} \\
\cline{4-6} \cline{7-13}
&&& \multicolumn{1}{c}{WikiText2} & 
\multicolumn{1}{c}{Ptb} & 
\multicolumn{1}{c}{C4} &
\multicolumn{1}{c}{BoolQ} & 
\multicolumn{1}{c}{HellaSwag} & 
\multicolumn{1}{c}{PIQA} & 
\multicolumn{1}{c}{WinoGrande} & 
\multicolumn{1}{c}{ARC-c} & 
\multicolumn{1}{c}{ARC-e} & 
\multicolumn{1}{c}{Avg.} \\
% \addlinespace[1ex]

\midrule
\multirow{7}{*}{LLaMA-7B}
% \rowcolor[gray]{0.9} % 设置第一行颜色，0=黑，1=白，0.9是浅灰色
                          & FP\cellcolor{gray!30}& 16\cellcolor{gray!30} & \cellcolor{gray!30}5.68&\cellcolor{gray!30}41.15&\cellcolor{gray!30}7.34&75.11\cellcolor{gray!30}  & 56.94\cellcolor{gray!30} & 78.67\cellcolor{gray!30}&70.01\cellcolor{gray!30}&41.89\cellcolor{gray!30}&75.25\cellcolor{gray!30}&66.31 \cellcolor{gray!30} \\
                          & GPTQ  & 2  &3.164e3&2.86e4&7.72e4& 45.47&25.85&52.01&48.30&23.55&25.42&36.77\\
                          & AWQ  & 2 &2.60e5&2.78e5&2.88e5& 37.83&25.28&52.72&49.25&22.44&25.25&35.46\\
                          & QuIP  & 2 &21.22&231.06&20.02& 52.94&36.93&62.51&55.41&23.04&40.45&45.21\\
                           & OmniQuant  & 2 & \cellcolor{red!10}9.23&\cellcolor{red!10}93.7&\cellcolor{red!10}12.1&\cellcolor{blue!10}64.80&\cellcolor{blue!10}42.52&\cellcolor{blue!10}69.53&56.35&\cellcolor{blue!10}27.65&\cellcolor{blue!10}60.65&\cellcolor{blue!10}53.58\\
                          & PB-LLM  & 1.6 & 12.45&269.73&27.49&62.69&34.05&61.10&\cellcolor{blue!10}57.38&22.18&45.58&47.16\\
        
                          & Squeeze10-LLM  & 1.6 & \cellcolor{blue!10}9.73&\cellcolor{blue!10}94.32&\cellcolor{blue!10}12.38&\cellcolor{red!10}65.02&\cellcolor{red!10}46.09&\cellcolor{red!10}72.91&\cellcolor{red!10}60.77&\cellcolor{red!10}34.64&\cellcolor{red!10}64.44&\cellcolor{red!10}\textbf{57.31}\\
\midrule
\multirow{7}{*}{LLaMA-30B} & FP\cellcolor{gray!30} & 16\cellcolor{gray!30} & \cellcolor{gray!30}4.1&\cellcolor{gray!30}23.51&\cellcolor{gray!30}6.13&82.69\cellcolor{gray!30}&63.33\cellcolor{gray!30}&80.96\cellcolor{gray!30}&76.01\cellcolor{gray!30}&52.82\cellcolor{gray!30}&80.43\cellcolor{gray!30}&72.71\cellcolor{gray!30} \\
                          & GPTQ  & 2 &161.33&1.12e4&8.61e3& 38.59&26.24&51.80&47.99&22.10&27.02&35.62\\
                          & AWQ  & 2 & 2.35e5&2.21e5&2.39e5&62.17&25.37&52.77&48.86&23.46&24.79&39.57\\
                          & QuIP  & 2 & 8.26&31.65&9.64& 66.76&49.74&73.34&64.25&31.74&67.26&58.85\\
                           & OmniQuant  & 2 & 7.14&\cellcolor{red!10}26.46&\cellcolor{red!10}9.1&66.76&\cellcolor{red!10}53.35&\cellcolor{blue!10}74.48&66.61&37.63&\cellcolor{blue!10}72.18&61.84\\
                          & PB-LLM  & 1.6 &\cellcolor{blue!10}6.74&41.58&12.52&\cellcolor{red!10}71.35&49.67&73.61&\cellcolor{red!10}72.77&\cellcolor{blue!10}39.42&71.59&\cellcolor{blue!10}63.07\\
                          & Squeeze10-LLM  & 1.6 &  \cellcolor{red!10}6.55&\cellcolor{blue!10}41.09&\cellcolor{blue!10}9.12&\cellcolor{blue!10}70.00&\cellcolor{blue!10}51.65&\cellcolor{red!10}77.20&\cellcolor{blue!10}69.93&\cellcolor{red!10}42.75&\cellcolor{red!10}74.37&\cellcolor{red!10}\textbf{64.32}\\
\midrule
\multirow{7}{*}{LLaMA-65B} & FP\cellcolor{gray!30}  & 16\cellcolor{gray!30} & \cellcolor{gray!30}3.53&\cellcolor{gray!30}25.07&\cellcolor{gray!30}5.81&84.89\cellcolor{gray!30}&64.56\cellcolor{gray!30}&81.28\cellcolor{gray!30}&77.35\cellcolor{gray!30}&52.82\cellcolor{gray!30}&81.31\cellcolor{gray!30}&73.7\cellcolor{gray!30}\\
                          & GPTQ  & 2  & 27.25&413.92&98.78&41.65&27.10&53.26&49.64&22.35&27.31&36.89\\
                          & AWQ  & 2  &7.39e4&6.80e4&7.51e4& 37.83&25.48&53.21&49.25&22.35&25.08&35.53\\
                          & QuIP  & 2  &6.8&\cellcolor{red!10}30.47&8.28& 76.54&53.92&76.55&69.61&37.29&69.87&63.96\\
                           & OmniQuant  & 2  &5.65&--&\cellcolor{red!10}7.60 &64.43&39.06&62.08&52.01&24.06&49.16&48.47\\
                          & PB-LLM  & 1.6  & \cellcolor{blue!10}5.43&48.94&9.81&\cellcolor{red!10}84.25&\cellcolor{blue!10}55.17&\cellcolor{blue!10}76.71&\cellcolor{red!10}74.35&\cellcolor{blue!10}44.37&\cellcolor{blue!10}77.86&\cellcolor{blue!10}68.79\\
                          & Squeeze10-LLM  & 1.6  & \cellcolor{red!10}5.34&\cellcolor{blue!10}31.31&\cellcolor{blue!10}7.72&\cellcolor{blue!10}79.48&\cellcolor{red!10}60.31&\cellcolor{red!10}79.60&\cellcolor{blue!10}73.48&\cellcolor{red!10}49.40&\cellcolor{red!10}79.55&\cellcolor{red!10}\textbf{70.30}\\
\bottomrule
\end{tabular}
}
\end{small}
% \end{center}
% \vskip -0.1in
\end{table*}

\subsection{Comparison with the State-of-the-Arts}
We have carried out quantitative experiments on three different sizes of models from two generations of the LLaMA family. Tables\,\ref{Llama2 result} and \ref{Llama1 result} provide a performance comparison of the LLaMA2 models (7B-70B) and LLaMA models (7B-65B) on six zero-shot classification tasks and three text-generation tasks, as well as the average bit count across various methods. Our proposed framework achieves the best results on all six models while maintaining an average bit count that is comparable to or even lower than other weight-only quantization methods and partially binarized methods.

As shown in Table\,\ref{Llama2 result}, on the LLaMA2-7B and LLaMA2-13B models, our method outperforms all other methods by 10-20\% across the six datasets under an extremely low bit count (i.e., 1.6 bits).
As the model size increases, the sparsity becomes more pronounced. Nevertheless, on the LLaMA2-70B model, our method surpasses the state-of-the-art post-training quantization (PTQ) methods (i.e., PB-LLM and OmniQuant) by an average of 3.1\%, and even almost matches the performance of the full-precision (FP) model, with only a 2.93\% accuracy loss. Also, as shown in Table\,\ref{Llama1 result}, in the first-generation LLaMA models, our method ranks among the top two across all datasets and achieves the highest average accuracy. For example, on the LLaMA-65B model, our method outperforms PB-LLM by an average of 1.5\%, and achieves a compression ratio of 10$\times$, with only a 3.4\% accuracy loss compared to the FP model.

\subsection{Analysis of Staged Quantization\label{subsec:analysis}}
In Figure\,\ref{fig:Multi-Bit-Compare}, we illustrate the relationship between the activation distribution of the original pretrained LLaMA2-7B model and those of its quantized counterparts, obtained by combining binarization with various high-bit quantization levels (ranging from 2-bit to 8-bit). Notably, as the bit-width increases, the activation distribution of the quantized model becomes less concentrated and more dispersed, with the 4-bit setting exhibiting the closest resemblance to the full-precision distribution. This observation is further supported by the KL divergence, which quantifies the discrepancy between the activation distributions of the quantized and original models.

% 比较2-8比特
\begin{table}[!t]
% \caption{distributions of different bits}
% \label{different bits anaysis}
\renewcommand{\arraystretch}{1.0} % 调整行高
\vskip 0.15in
\begin{center}
\caption{The quantized LLaMA2-7B model accuracy obtained when using different bit-width as intermediate bits.}
\label{different bits anaysis}
%\vskip -0.1in
%\begin{small}
\resizebox{0.48\textwidth}{!}{
\begin{tabular}{cccccc}
\toprule
Bit-Width & BoolQ & HellaSwag & PIQA & WinoGrange \\
\midrule
2   & 43.79 & 25.43 & 52.77 & 50.12 \\
3   & 47.61 & 32.67 & 58.76 & 52.01 \\ 
\textbf{4}   & \textbf{66.54} & \textbf{46.03} & \textbf{72.20} & \textbf{64.25} \\
5   & 63.91 & 37.60 & 67.36 & 57.46 \\
6   & 61.04 & 26.37 & 54.90 & 49.09 \\
7   & 37.83 & 25.08 & 49.51 & 48.78 \\
8   & 37.83 & 25.09 & 49.51 & 50.12 \\

\bottomrule
\end{tabular}
}
%\end{small}
\end{center}
\vskip -0.2in
\end{table}

\begin{table*}[!t]
\caption{Performance comparison of different proportion of high-bit (4bit) on LLaMA2-70B on three text-generation tasks and six zero-shot classification tasks.}
\label{different mean-bit compare}
%rop\vskip -0.1in
% \begin{center}
\renewcommand{\arraystretch}{1.2} % 调整行高
\begin{small}
\resizebox{\linewidth}{!}{  % 使表格宽度适应页面横宽
\begin{tabular}{cc|ccc|cccccc|cc}
\toprule
% \cline{4-10}  % 在accuracy下画线
% \multicolumn{1}{c}{\textbf{Model}} & 
% \multicolumn{1}{c}{\textbf{Method}} & 
% \multicolumn{1}{c}{\textbf{WBits}} &
% \multirow{2}{*}{\textbf{Model}} & 
\multirow{2}{*}{\textbf{High Proportion}} & 
\multirow{2}{*}{\textbf{\#W-Bits}} & 
\multicolumn{3}{c}{\textbf{Perplexity$\downarrow$}} &
\multicolumn{7}{c}{\textbf{Accuracy(\%)$\uparrow$}} \\
\cline{3-5} \cline{6-12}
&& \multicolumn{1}{c}{WikiText2} & 
\multicolumn{1}{c}{Ptb} & 
\multicolumn{1}{c}{C4} &
\multicolumn{1}{c}{BoolQ} & 
\multicolumn{1}{c}{HellaSwag} & 
\multicolumn{1}{c}{PIQA} & 
\multicolumn{1}{c}{WinoGrande} & 
\multicolumn{1}{c}{ARC-c} & 
\multicolumn{1}{c}{ARC-e} & 
\multicolumn{1}{c}{Avg.} \\
% \addlinespace[1ex]

\midrule
% \multirow{7}{*}{LLaMA2-70B}
% \rowcolor[gray]{0.9} % 设置第一行颜色，0=黑，1=白，0.9是浅灰色
                          \cellcolor{gray!30}FP  & \cellcolor{gray!30}16  & \cellcolor{gray!30}3.32&\cellcolor{gray!30}24.25&\cellcolor{gray!30}5.71&\cellcolor{gray!30}83.79&\cellcolor{gray!30}64.77&\cellcolor{gray!30}82.21&\cellcolor{gray!30}77.90&\cellcolor{gray!30}54.35&\cellcolor{gray!30}82.74 &\cellcolor{gray!30}74.29 \\
                          
                    60\%  & 2.8 & 4.01&25.08&6.31&81.22&61.53&81.23&76.40&51.88&79.92&72.03\\ 

                        50\%  & 2.5 & 4.13&25.61&6.43&80.12&60.97&80.41&75.45&51.88&79.84&71.45\\

                        40\%  & 2.2 &4.22&25.83&6.52&79.39&61.05&80.3&74.35&50.85&79.67&70.94\\

                        30\%  & 1.9& 4.36&26.29&6.67&79.27&60.88&81.07&75.30&49.49&79.42&70.91\\
                       
                         \textbf{20\%(Ours)}  & \textbf{1.6}  &4.74&28.31&7.11&80.76&60.01&79.38&76.87&49.15&78.91&70.85\\

                           10\%  & 1.3  &7.05&70.04&10.55&80.98 &54.57 &77.15 &74.11 &46.08 &74.96 &67.98 \\

\bottomrule
\end{tabular}
}
\end{small}
% \end{center}
% \vskip -0.1in
\end{table*}

Additionally, Table\,\ref{different bits anaysis} presents the impact of mixing binarization with different high-bit quantization levels on LLaMA2-7B. Specifically, we examine model accuracy under the condition that 20\% of the salient weights are retained within the 2-bit to 8-bit range. Interestingly, we find that, given the same proportion of retained salient weights, the combination of binarization and 4-bit quantization yields the highest performance. This result is somewhat counterintuitive, as higher bit-widths (5-bit to 8-bit) theoretically preserve more original weight information. However, our findings suggest that 4-bit quantization strikes the optimal balance between information retention and quantization efficiency.
We speculate that 4-bit serves as an effective intermediate representation, striking a balance between the need for higher precision and the significant gap between 1-bit and higher-bit configurations.

% Furthermore, the experimental results in Table \ref{different bits anaysis} indicate that, despite the theoretical advantages of higher bit-widths, the 4-bit configuration consistently outperforms other settings. We speculate that 4-bit serves as an effective intermediate representation, striking a balance between the need for higher precision and the significant gap between 1-bit and higher-bit configurations.

% Table\,\ref{different bits anaysis} reports the effect of mixing binarization with high-bit quantization (2-bit to 8-bit). 
% We compare the model accuracy when the salient weights were in the range of 2-8 bits, with a proportion of 0.2. We found that, for the same proportion of salient weights retained, binarization and 4-bit quantization yielded the best performance, despite the fact that higher bit-widths(5-bit to 8-bit) theoretically retain more original weight information. Furthermore, ~{\color{yellow}Figure 5} illustrates that 4-bit quantization combined with binarization results in an activation distribution that is closest to the activation distribution of the original model.

\subsection{Analysis of Salient Weight Proportion}
Table\,\ref {different mean-bit compare} examines the impact of salient weight proportions on model performance. Note that salient weights and non-salient weights are quantized to 4-bit and 1-bit respectively. We compare the performance of the quantized model on six zero-shot classification tasks and three perplexity tasks. Clearly, a higher proportion of salient weights leads to better performance.
Moreover, we observe that when the mean-bit ranges from 1.6 bit to 2.8 bit, the performance of the quantized model remains relatively close to FP. Specifically, the best accuracy (2.8-bit) is only 2.26\% lower than FP, while the worst (1.6-bit) is 3.44\% lower than FP. However, when the proportion of salient weights is only 10\%, the performance of the quantized model significantly deteriorates across all aspects. To meet the requirements of ultra-low-bit quantization, we select a 20\% proportion of salient weights, leading to an average 1.6-bit quantized model. 
% Of course, other proportions can also be chosen, but we do not further investigate them in this paper.

% % 消融实验,技术对比
% \begin{table}[htbp]
% % \caption{Ablation Study of tricks}
% % \label{Ablation Study of tricks}
% %\vskip 0.15in
% \begin{center}
% \caption{Effects of PBAR and FIAS.}
% \label{Ablation Study of tricks}
% %\vskip -0.1in
% \begin{small}
% \begin{tabular}{lcc}
% \toprule
%         \textbf{Method} & \textbf{WikiText2$\downarrow$} & \textbf{WinoGrange$\uparrow$} \\
% \midrule
% % PB-LLM     & 12.29  & 56.43\% \\
% % \hline
% \rowcolor[gray]{0.9} % 设置第一行颜色，0=黑，1=白，0.9是浅灰色
% \textbf{Ours}          & \textbf{9.96}  & \textbf{64.56\%} \\
% -PBAR        & 10.01  & 64.33\% \\
% -FIAS       & 9.97 & 62.19\% \\
% -PBAR-FIAS   & 10.16  & 61.25\% \\
% % PB-LLM     & 12.29  & 56.43\% \\
% \bottomrule
% \end{tabular}
% \end{small}
% \end{center}
% \vskip -0.2in
% \end{table}

% 消融实验,技术对比
\begin{table}[htbp]
% \caption{Ablation Study of tricks}
% \label{Ablation Study of tricks}
%\vskip 0.15in
\begin{center}
\caption{Effects of PBAR and FIAS.}
\label{Ablation Study of tricks}
%\vskip -0.1in
\begin{small}
\begin{tabular}{lcc}
\toprule
        \textbf{Method} & \textbf{WinoGrange$\uparrow$} \\
\midrule
% PB-LLM     & 12.29  & 56.43\% \\
% \hline
\rowcolor[gray]{0.9} % 设置第一行颜色，0=黑，1=白，0.9是浅灰色
\textbf{Ours}      & \textbf{64.56\%} \\
-PBAR     & 64.33\% \\
-FIAS     & 62.19\% \\
-PBAR-FIAS  & 61.25\% \\
-PBAR-FIAS-Staged Quantization  & 56.43\% \\
\bottomrule
\end{tabular}
\end{small}
\end{center}
\vskip -0.2in
\end{table}

\subsection{Ablation Study}

\noindent \textbf{Effects of PBAR and FIAS.} In Table\,\ref{Ablation Study of tricks}, we ablate the impact of PBAR and FIAS by replacing
them separately, and show performance changes in perplexity (WikiText2) and accuracy (WinoGrande). 
% \tr{Due to the usage of 4 bit intead of XXX-bit of PB-LLM, we can see that ours significantly outperform the baseline method PB-LLM.} 
Specifically, we replace PBAR and FIAS with standard practices, Hessian-based weight salience measurement~\cite{frantar2023sparsegpt,shang2023pb} and quantized activation information~\cite{chee2023quip,shao2023omniquant,frantar2022gptq,lin2024awq}, respectively.
% respectively means not using our improved methods in Sec.~\ref{sec:pbar} and Sec.~\ref{sec:fias}, but instead adopting Hessian-based weight significance measurement~\cite{frantar2023sparsegpt,shang2023pb} or using quantized activation information~\cite{chee2023quip,shao2023omniquant,frantar2022gptq,lin2024awq}.
{Replacing PBAR (i.e., ``-PBAR'') leads to a 0.05 increase in perplexity (WikiText2) and a 0.23\% decrease in accuracy (WinoGrande). Replacing PBAR (i.e., ``-FIAS'') leads to a 0.01 increase in perplexity (WikiText2) and a 2.37\% decrease in accuracy (WinoGrande). Also, replacing both further worsens the results.}
% ``-PBAR'' indicates that PBAR is not used. The combination of PBAR and FIAS leads to a 3.31\% improvement in accuracy, further demonstrating the superiority of our framework.

\noindent \textbf{Hyperparameter Analysis of $\lambda$.} We analyzed the selection of hyperparameter $\lambda$ in Appendix \ref{appendix-lamabda} and Table\,\ref{lambda analysis}. 

\section{Conclusion}
\label{sec:conclusion}

In this paper, we have proposed Squeeze10-LLM, a mixed-precision ultra-low bit post-training quantization method, to balance model compression ratios and performance degradation. Building on the intrinsic correlation between activation value ranges and representational capacity, we introduce Quantization with Activation Robustness (PBAR) to refine the weight salience metric and establish a systematic 4-bit allocation strategy. Furthermore, by analyzing the interdependence mechanism between activations and quantization, we introduce Full Information Activation Supervision (FIAS) to mitigate progressive distributional shifts across layers. Extensive experimental results show that our proposed Squeeze10-LLM outperforms other $\le$ 2-bit state-of-the-arts that are particularly designed for LLMs' quantization.

% \input{sec/7_Acknowledgement}

% \newpage
{
    \small
    \bibliographystyle{plain}
    \bibliography{references}

@article{frantar2022gptq,
  title={Gptq: Accurate post-training quantization for generative pre-trained transformers},
  author={Frantar, Elias and Ashkboos, Saleh and Hoefler, Torsten and Alistarh, Dan},
  journal={arXiv preprint arXiv:2210.17323},
  year={2022}
}

@article{frantar2022optimal,
  title={Optimal brain compression: A framework for accurate post-training quantization and pruning},
  author={Frantar, Elias and Alistarh, Dan},
  journal={Advances in Neural Information Processing Systems},
  volume={35},
  pages={4475--4488},
  year={2022}
}

@article{chee2023quip,
  title={Quip: 2-bit quantization of large language models with guarantees},
  author={Chee, Jerry and Cai, Yaohui and Kuleshov, Volodymyr and De Sa, Christopher M},
  journal={Advances in Neural Information Processing Systems},
  volume={36},
  pages={4396--4429},
  year={2023}
}

@article{lin2024awq,
  title={Awq: Activation-aware weight quantization for on-device llm compression and acceleration},
  author={Lin, Ji and Tang, Jiaming and Tang, Haotian and Yang, Shang and Chen, Wei-Ming and Wang, Wei-Chen and Xiao, Guangxuan and Dang, Xingyu and Gan, Chuang and Han, Song},
  journal={Proceedings of Machine Learning and Systems},
  volume={6},
  pages={87--100},
  year={2024}
}

@inproceedings{lee2024owq,
  title={Owq: Outlier-aware weight quantization for efficient fine-tuning and inference of large language models},
  author={Lee, Changhun and Jin, Jungyu and Kim, Taesu and Kim, Hyungjun and Park, Eunhyeok},
  booktitle={Proceedings of the AAAI Conference on Artificial Intelligence},
  volume={38},
  number={12},
  pages={13355--13364},
  year={2024}
}

@inproceedings{xiao2023smoothquant,
  title={Smoothquant: Accurate and efficient post-training quantization for large language models},
  author={Xiao, Guangxuan and Lin, Ji and Seznec, Mickael and Wu, Hao and Demouth, Julien and Han, Song},
  booktitle={International Conference on Machine Learning},
  pages={38087--38099},
  year={2023},
  organization={PMLR}
}

@article{wei2022outlier,
  title={Outlier suppression: Pushing the limit of low-bit transformer language models},
  author={Wei, Xiuying and Zhang, Yunchen and Zhang, Xiangguo and Gong, Ruihao and Zhang, Shanghang and Zhang, Qi and Yu, Fengwei and Liu, Xianglong},
  journal={Advances in Neural Information Processing Systems},
  volume={35},
  pages={17402--17414},
  year={2022}
}

@article{shao2023omniquant,
  title={Omniquant: Omnidirectionally calibrated quantization for large language models},
  author={Shao, Wenqi and Chen, Mengzhao and Zhang, Zhaoyang and Xu, Peng and Zhao, Lirui and Li, Zhiqian and Zhang, Kaipeng and Gao, Peng and Qiao, Yu and Luo, Ping},
  journal={arXiv preprint arXiv:2308.13137},
  year={2023}
}

@article{yao2022zeroquant,
  title={Zeroquant: Efficient and affordable post-training quantization for large-scale transformers},
  author={Yao, Zhewei and Yazdani Aminabadi, Reza and Zhang, Minjia and Wu, Xiaoxia and Li, Conglong and He, Yuxiong},
  journal={Advances in Neural Information Processing Systems},
  volume={35},
  pages={27168--27183},
  year={2022}
}

@article{yuan2023rptq,
  title={Rptq: Reorder-based post-training quantization for large language models},
  author={Yuan, Zhihang and Niu, Lin and Liu, Jiawei and Liu, Wenyu and Wang, Xinggang and Shang, Yuzhang and Sun, Guangyu and Wu, Qiang and Wu, Jiaxiang and Wu, Bingzhe},
  journal={arXiv preprint arXiv:2304.01089},
  year={2023}
}

@article{touvron2023llama1,
  title={Llama: Open and efficient foundation language models},
  author={Touvron, Hugo and Lavril, Thibaut and Izacard, Gautier and Martinet, Xavier and Lachaux, Marie-Anne and Lacroix, Timoth{\'e}e and Rozi{\`e}re, Baptiste and Goyal, Naman and Hambro, Eric and Azhar, Faisal and others},
  journal={arXiv preprint arXiv:2302.13971},
  year={2023}
}

@article{touvron2023llama2,
  title={Llama 2: Open foundation and fine-tuned chat models},
  author={Touvron, Hugo and Martin, Louis and Stone, Kevin and Albert, Peter and Almahairi, Amjad and Babaei, Yasmine and Bashlykov, Nikolay and Batra, Soumya and Bhargava, Prajjwal and Bhosale, Shruti and others},
  journal={arXiv preprint arXiv:2307.09288},
  year={2023}
}

@article{zellers2019hellaswag,
  title={Hellaswag: Can a machine really finish your sentence?},
  author={Zellers, Rowan and Holtzman, Ari and Bisk, Yonatan and Farhadi, Ali and Choi, Yejin},
  journal={arXiv preprint arXiv:1905.07830},
  year={2019}
}

@article{sakaguchi2021winogrande,
  title={Winogrande: An adversarial winograd schema challenge at scale},
  author={Sakaguchi, Keisuke and Bras, Ronan Le and Bhagavatula, Chandra and Choi, Yejin},
  journal={Communications of the ACM},
  volume={64},
  number={9},
  pages={99--106},
  year={2021},
  publisher={ACM New York, NY, USA}
}

@inproceedings{bisk2020piqa,
  title={Piqa: Reasoning about physical commonsense in natural language},
  author={Bisk, Yonatan and Zellers, Rowan and Gao, Jianfeng and Choi, Yejin and others},
  booktitle={Proceedings of the AAAI conference on artificial intelligence},
  volume={34},
  number={05},
  pages={7432--7439},
  year={2020}
}

@article{clark2019boolq,
  title={Boolq: Exploring the surprising difficulty of natural yes/no questions},
  author={Clark, Christopher and Lee, Kenton and Chang, Ming-Wei and Kwiatkowski, Tom and Collins, Michael and Toutanova, Kristina},
  journal={arXiv preprint arXiv:1905.10044},
  year={2019}
}

@article{clark2018think,
  title={Think you have solved question answering? try arc, the ai2 reasoning challenge},
  author={Clark, Peter and Cowhey, Isaac and Etzioni, Oren and Khot, Tushar and Sabharwal, Ashish and Schoenick, Carissa and Tafjord, Oyvind},
  journal={arXiv preprint arXiv:1803.05457},
  year={2018}
}

@article{merity2016pointer,
  title={Pointer sentinel mixture models},
  author={Merity, Stephen and Xiong, Caiming and Bradbury, James and Socher, Richard},
  journal={arXiv preprint arXiv:1609.07843},
  year={2016}
}

@article{raffel2020exploring,
  title={Exploring the limits of transfer learning with a unified text-to-text transformer},
  author={Raffel, Colin and Shazeer, Noam and Roberts, Adam and Lee, Katherine and Narang, Sharan and Matena, Michael and Zhou, Yanqi and Li, Wei and Liu, Peter J},
  journal={Journal of machine learning research},
  volume={21},
  number={140},
  pages={1--67},
  year={2020}
}

@article{marcus-etal-1993-building, 
    title = "Building a Large Annotated Corpus of {E}nglish: The {P}enn {T}reebank", 
    author = "Marcus, Mitchell P. and Santorini, Beatrice and Marcinkiewicz, Mary Ann",
    journal = "Computational Linguistics",
    volume = "19",
    number = "2",
    year = "1993",
    url = "https://www.aclweb.org/anthology/J93-2004",
    pages = "313--330"
}

@article{shang2023pb,
  title={Pb-llm: Partially binarized large language models},
  author={Shang, Yuzhang and Yuan, Zhihang and Wu, Qiang and Dong, Zhen},
  journal={arXiv preprint arXiv:2310.00034},
  year={2023}
}

@article{zheng2024mixllm,
  title={MixLLM: LLM Quantization with Global Mixed-precision between Output-features and Highly-efficient System Design},
  author={Zheng, Zhen and Song, Xiaonan and Liu, Chuanjie},
  journal={arXiv preprint arXiv:2412.14590},
  year={2024}
}

@inproceedings{li2023llm,
  title={Llm-mq: Mixed-precision quantization for efficient llm deployment},
  author={Li, Shiyao and Ning, Xuefei and Hong, Ke and Liu, Tengxuan and Wang, Luning and Li, Xiuhong and Zhong, Kai and Dai, Guohao and Yang, Huazhong and Wang, Yu},
  booktitle={NeurIPS 2023 Efficient Natural Language and Speech Processing Workshop},
  pages={1--5},
  year={2023}
}

@article{huang2024slim,
  title={SliM-LLM: Salience-driven mixed-precision quantization for large language models},
  author={Huang, Wei and Qin, Haotong and Liu, Yangdong and Li, Yawei and Liu, Xianglong and Benini, Luca and Magno, Michele and Qi, Xiaojuan},
  journal={arXiv preprint arXiv:2405.14917},
  year={2024}
}

@article{chen2024progressive,
  title={Progressive Mixed-Precision Decoding for Efficient LLM Inference},
  author={Chen, Hao Mark and Tan, Fuwen and Kouris, Alexandros and Lee, Royson and Fan, Hongxiang and Venieris, Stylianos I},
  journal={arXiv preprint arXiv:2410.13461},
  year={2024}
}

@inproceedings{guan2024aptq,
  title={Aptq: Attention-aware post-training mixed-precision quantization for large language models},
  author={Guan, Ziyi and Huang, Hantao and Su, Yupeng and Huang, Hong and Wong, Ngai and Yu, Hao},
  booktitle={Proceedings of the 61st ACM/IEEE Design Automation Conference},
  pages={1--6},
  year={2024}
}

@inproceedings{ou2024adaptive,
  title={Adaptive quantization error reconstruction for llms with mixed precision},
  author={Ou, Lin and Xia, Jinpeng and Zhang, Yuewei and Hao, Chuzhan and Wang, Hao Henry},
  booktitle={First Conference on Language Modeling},
  year={2024}
}

@inproceedings{guo2024compressing,
  title={Compressing large language models by joint sparsification and quantization},
  author={Guo, Jinyang and Wu, Jianyu and Wang, Zining and Liu, Jiaheng and Yang, Ge and Ding, Yifu and Gong, Ruihao and Qin, Haotong and Liu, Xianglong},
  booktitle={Forty-first International Conference on Machine Learning},
  year={2024}
}

@inproceedings{frantar2023sparsegpt,
  title={Sparsegpt: Massive language models can be accurately pruned in one-shot},
  author={Frantar, Elias and Alistarh, Dan},
  booktitle={International Conference on Machine Learning},
  pages={10323--10337},
  year={2023},
  organization={PMLR}
}

@article{liu2024deepseek-v3,
  title={Deepseek-v3 technical report},
  author={Liu, Aixin and Feng, Bei and Xue, Bing and Wang, Bingxuan and Wu, Bochao and Lu, Chengda and Zhao, Chenggang and Deng, Chengqi and Zhang, Chenyu and Ruan, Chong and others},
  journal={arXiv preprint arXiv:2412.19437},
  year={2024}
}

@article{guo2025deepseek,
  title={Deepseek-r1: Incentivizing reasoning capability in llms via reinforcement learning},
  author={Guo, Daya and Yang, Dejian and Zhang, Haowei and Song, Junxiao and Zhang, Ruoyu and Xu, Runxin and Zhu, Qihao and Ma, Shirong and Wang, Peiyi and Bi, Xiao and others},
  journal={arXiv preprint arXiv:2501.12948},
  year={2025}
}

@article{liu2024deepseek-v2,
  title={Deepseek-v2: A strong, economical, and efficient mixture-of-experts language model},
  author={Liu, Aixin and Feng, Bei and Wang, Bin and Wang, Bingxuan and Liu, Bo and Zhao, Chenggang and Dengr, Chengqi and Ruan, Chong and Dai, Damai and Guo, Daya and others},
  journal={arXiv preprint arXiv:2405.04434},
  year={2024}
}

@article{lu2024deepseek,
  title={Deepseek-vl: towards real-world vision-language understanding},
  author={Lu, Haoyu and Liu, Wen and Zhang, Bo and Wang, Bingxuan and Dong, Kai and Liu, Bo and Sun, Jingxiang and Ren, Tongzheng and Li, Zhuoshu and Yang, Hao and others},
  journal={arXiv preprint arXiv:2403.05525},
  year={2024}
}

@article{achiam2023gpt,
  title={Gpt-4 technical report},
  author={Achiam, Josh and Adler, Steven and Agarwal, Sandhini and Ahmad, Lama and Akkaya, Ilge and Aleman, Florencia Leoni and Almeida, Diogo and Altenschmidt, Janko and Altman, Sam and Anadkat, Shyamal and others},
  journal={arXiv preprint arXiv:2303.08774},
  year={2023}
}

@article{hurst2024gpt,
  title={Gpt-4o system card},
  author={Hurst, Aaron and Lerer, Adam and Goucher, Adam P and Perelman, Adam and Ramesh, Aditya and Clark, Aidan and Ostrow, AJ and Welihinda, Akila and Hayes, Alan and Radford, Alec and others},
  journal={arXiv preprint arXiv:2410.21276},
  year={2024}
}

@article{jaech2024openai,
  title={Openai o1 system card},
  author={Jaech, Aaron and Kalai, Adam and Lerer, Adam and Richardson, Adam and El-Kishky, Ahmed and Low, Aiden and Helyar, Alec and Madry, Aleksander and Beutel, Alex and Carney, Alex and others},
  journal={arXiv preprint arXiv:2412.16720},
  year={2024}
}
}

\newpage
\appendix
\section{Appendix}
\label{sec:Appendix}
% \subsection{A.Selection of high-bit quantization bit-width}
% In our mixed bit quantization framework, we discussed the selection of salient weights with different bit-widths (ranging from 2 to 8 bits) and their combination with binarization for mixed quantization. As shown in Table 7, we compared the model accuracy when the significant weights were in the range of 2-8 bits, with a proportion of 0.2. We found that, for the same proportion of salient weights retained, binarization and 4-bit quantization yielded the best performance, despite the higher bit-widths (5-8 bits) retaining more original weight information.

% \input{tabs/tabel6}

\subsection{The analysis of selecting high bit precision}
\label{appendix-high-bit}
\begin{figure}[htbp]
    \centering
    \includegraphics[width=0.49\textwidth,keepaspectratio]{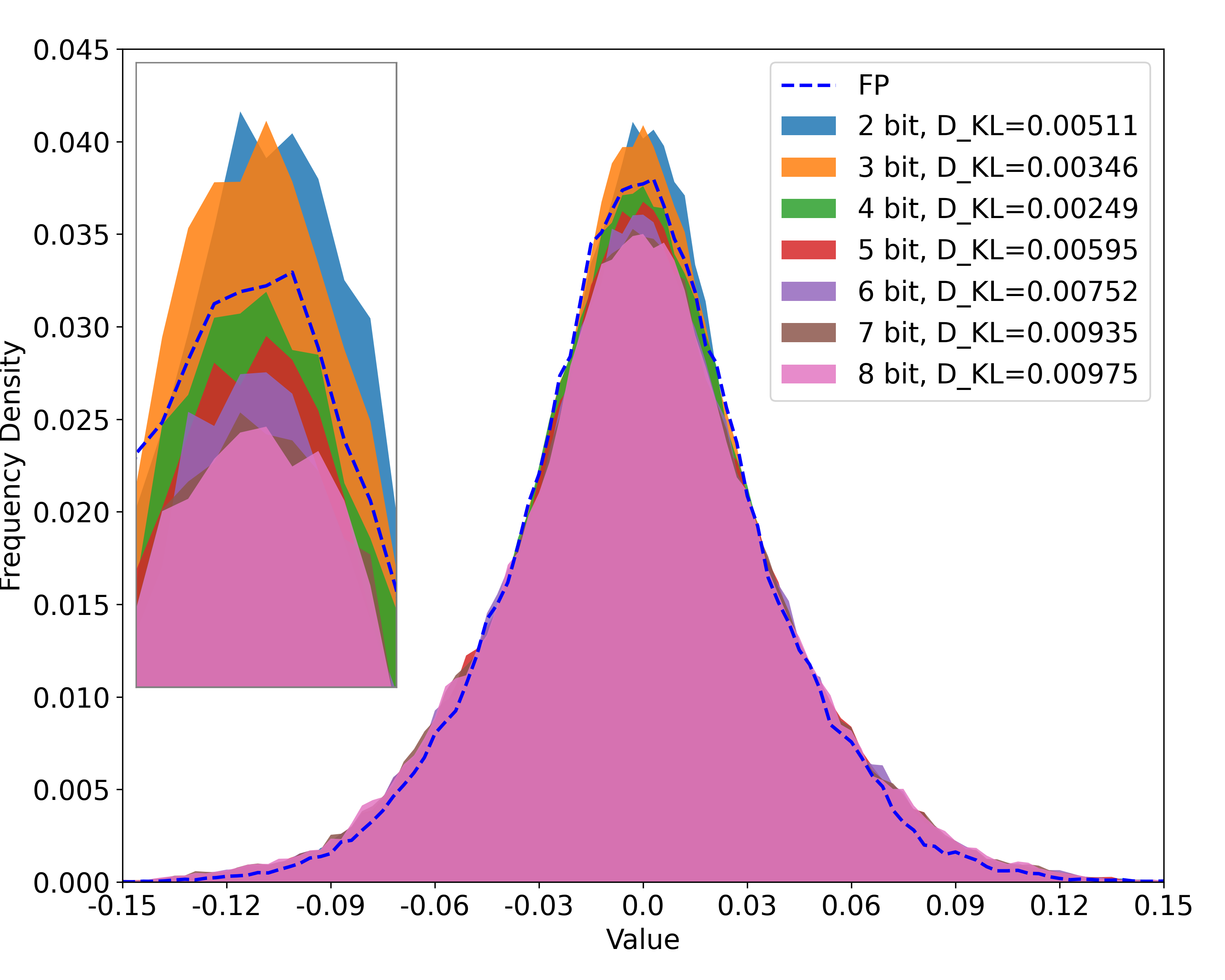}
    \caption{Comparison of the frequency density distributions of the activation output from the output projection of the 6-th layer in LLaMA2-7b when applying 2 to 8 bits as the intermediate bitwidths in Staged Mixed-Precision Quantization. Each \(D_{KL}\) indicates the Kullback-Leibler divergence between current activation value density distribution and its full precision (FP) counterpart. {It can be seen that when 4-bit is used, the distribution characteristics are the closest to those of the full-precision results.}}
    \label{fig:Multi-Bit-Compare}
\end{figure}

For intermediate bits, possible bits range from 2-bit to 8-bit. Figure\,\ref{fig:Multi-Bit-Compare} demonstrates that the activation distribution obtained using 4-bit quantization is the closest to that of the full precision (FP) model. The advantage of 4-bit as an intermediate representation is also confirmed by our experiments {(see Sec.\,\ref{subsec:analysis})}. Thus, we use 4-bit quantization.

\subsection[Hyperparameter Analysis of lambda]{Hyperparameter Analysis of $\lambda$}
\label{appendix-lamabda}
Table\,\ref{lambda analysis} analyzes the perplexity on WikiText2 datasets across different values of $\lambda$ of Eq.\,(\ref{Salience metric}).
For LLaMA2-7B, the best performance is achieved when $\lambda=3e-4$. This value is also adopted for the quantization of other models.
\begin{table}[ht]
% \caption{Analysis of Hyperparameter $\lambda$ on LLaMa-2-7b}
% \label{lambda analysis}
%\vskip 0.2in
\begin{center}
\caption{Analysis of hyperparameter $\lambda$ on LLaMA2-7b.}
\label{lambda analysis}
%\vskip -0.1in
\begin{small}
% \begin{tabular}{0.5\linewidth}{lccccr}
\begin{tabular}{cccccc}
\toprule
$\lambda$ & 1e-2 & 1e-3 & 3e-4 & 1e-4 & 1e-5  \\
\midrule
WikiText2 & 14.03  & 9.99 & \textbf{9.96} & 10.01 & 10.10\\
\bottomrule
\end{tabular}
\end{small}
\end{center}
\vskip -0.2in
\end{table}

\subsection{Salient Weight Storing Cost}
\label{appendix-lamabda}
The additional overhead for just storing the salient weights is acceptable. The  overall bit number, $\mathbf{N}_{bit}$ must adhere to the following condition:
\begin{equation}
    \mathbf{N}_{bit}\le 1\times r_{binary}+4\times(1-r_{binary})+1,
\end{equation}
where $r_{binary}$ denotes the ratio of the binarized weights, taking the value of 0.2. The additional 1 bit is allocated for index storage of salient weight, and the storage representation could be further optimized using sparse matrix storage methods such as Compressed Sparse Row.

\newpage
\section*{NeurIPS Paper Checklist}

%%% BEGIN INSTRUCTIONS %%%
%%% END INSTRUCTIONS %%%

\begin{enumerate}

\item {\bf Claims}
    \item[] Question: Do the main claims made in the abstract and introduction accurately reflect the paper's contributions and scope?
    \item[] Answer: \answerYes{} % Replace by \answerYes{}, \answerNo{}, or \answerNA{}.
    \item[] Justification: Our abstract and introduction include the three main techniques and experimental results presented in the paper.

\item {\bf Limitations}
    \item[] Question: Does the paper discuss the limitations of the work performed by the authors?
    \item[] Answer: \answerYes{} % Replace by \answerYes{}, \answerNo{}, or \answerNA{}.
    \item[] Justification: The appendix discusses the limitations of our method, including the additional memory overhead introduced by the mask.

\item {\bf Theory assumptions and proofs}
    \item[] Question: For each theoretical result, does the paper provide the full set of assumptions and a complete (and correct) proof?
    \item[] Answer: \answerNA{} % Replace by \answerYes{}, \answerNo{}, or \answerNA{}.
    \item[] Justification: We do not have theoretical contributions in this work, where our contributions
are validated with experiments.

    \item {\bf Experimental result reproducibility}
    \item[] Question: Does the paper fully disclose all the information needed to reproduce the main experimental results of the paper to the extent that it affects the main claims and/or conclusions of the paper (regardless of whether the code and data are provided or not)?
    \item[] Answer: \answerYes{} % Replace by \answerYes{}, \answerNo{}, or \answerNA{}.
    \item[] Justification: All datasets and models used are publicly available, and the quantization method includes all necessary implementation details.

\item {\bf Open access to data and code}
    \item[] Question: Does the paper provide open access to the data and code, with sufficient instructions to faithfully reproduce the main experimental results, as described in supplemental material?
    \item[] Answer: \answerYes{} % Replace by \answerYes{}, \answerNo{}, or \answerNA{}.
    \item[] Justification: The implementation of our method is not complex, and the core technical details have been disclosed in the paper. The code will be released soon.

\item {\bf Experimental setting/details}
    \item[] Question: Does the paper specify all the training and test details (e.g., data splits, hyperparameters, how they were chosen, type of optimizer, etc.) necessary to understand the results?
    \item[] Answer: \answerYes{} % Replace by \answerYes{}, \answerNo{}, or \answerNA{}.
    \item[] Justification: All parameter settings and experimental details necessary for reproduction are provided in the experimental section and the appendix.

\item {\bf Experiment statistical significance}
    \item[] Question: Does the paper report error bars suitably and correctly defined or other appropriate information about the statistical significance of the experiments?
    \item[] Answer: \answerYes{} % Replace by \answerYes{}, \answerNo{}, or \answerNA{}.
    \item[] Justification: The performance of all quantized models is evaluated using the authoritative open-source library for large language models, lm-eval (https://github.com/EleutherAI/lm-evaluation-harness). The results are fully reproducible by setting the random seed, with negligible variance.

\item {\bf Experiments compute resources}
    \item[] Question: For each experiment, does the paper provide sufficient information on the computer resources (type of compute workers, memory, time of execution) needed to reproduce the experiments?
    \item[] Answer: \answerYes{} % Replace by \answerYes{}, \answerNo{}, or \answerNA{}.
    \item[] Justification: We have provided that in the experimental section.
    
\item {\bf Code of ethics}
    \item[] Question: Does the research conducted in the paper conform, in every respect, with the NeurIPS Code of Ethics \url{https://neurips.cc/public/EthicsGuidelines}?
    \item[] Answer: \answerYes{} % Replace by \answerYes{}, \answerNo{}, or \answerNA{}.
    \item[] Justification: We have reviewed that and claim we conform that Code of Ethics.

\item {\bf Broader impacts}
    \item[] Question: Does the paper discuss both potential positive societal impacts and negative societal impacts of the work performed?
    \item[] Answer: \answerNA{} % Replace by \answerYes{}, \answerNo{}, or \answerNA{}.
    \item[] Justification: There are not direct paths to any negative applications.
    
\item {\bf Safeguards}
    \item[] Question: Does the paper describe safeguards that have been put in place for responsible release of data or models that have a high risk for misuse (e.g., pretrained language models, image generators, or scraped datasets)?
    \item[] Answer: \answerNA{} % Replace by \answerYes{}, \answerNo{}, or \answerNA{}.
    \item[] Justification: The paper does not have such risks.

\item {\bf Licenses for existing assets}
    \item[] Question: Are the creators or original owners of assets (e.g., code, data, models), used in the paper, properly credited and are the license and terms of use explicitly mentioned and properly respected?
    \item[] Answer: \answerYes{} % Replace by \answerYes{}, \answerNo{}, or \answerNA{}.
    \item[] Justification: We used widely available public datasets and have cited them in the references.

\item {\bf New assets}
    \item[] Question: Are new assets introduced in the paper well documented and is the documentation provided alongside the assets?
    % \item[] Answer: \answerNA{} % Replace by \answerYes{}, \answerNo{}, or \answerNA{}.
    % \item[] Justification: We use and cite existing datasets in this work. Other assets, including code/model, will be released after the publication.
    \item[] Answer: \answerNA{} % Replace by \answerYes{}, \answerNo{}, or \answerNA{}.
    \item[] Justification: No new assets.

\item {\bf Crowdsourcing and research with human subjects}
    \item[] Question: For crowdsourcing experiments and research with human subjects, does the paper include the full text of instructions given to participants and screenshots, if applicable, as well as details about compensation (if any)?
    \item[] Answer: \answerNA{} % Replace by \answerYes{}, \answerNo{}, or \answerNA{}.
    \item[] Justification: We do not include such experiments.

\item {\bf Institutional review board (IRB) approvals or equivalent for research with human subjects}
    \item[] Question: Does the paper describe potential risks incurred by study participants, whether such risks were disclosed to the subjects, and whether Institutional Review Board (IRB) approvals (or an equivalent approval/review based on the requirements of your country or institution) were obtained?
    \item[] Answer: \answerNA{} % Replace by \answerYes{}, \answerNo{}, or \answerNA{}.
    \item[] Justification: We do not include such experiments.

\item {\bf Declaration of LLM usage}
    \item[] Question: Does the paper describe the usage of LLMs if it is an important, original, or non-standard component of the core methods in this research? Note that if the LLM is used only for writing, editing, or formatting purposes and does not impact the core methodology, scientific rigorousness, or originality of the research, declaration is not required.
    %this research? 
    \item[] Answer: \answerYes{} % Replace by \answerYes{}, \answerNo{}, or \answerNA{}.
    \item[] Justification: The paper has described the usage of LLMs

\end{enumerate}

\end{document}